\documentclass{article}

\usepackage{amsmath,amssymb,amsthm,bbm,natbib,color,hyperref}
\usepackage[margin=1in]{geometry}
\usepackage{graphicx}
\usepackage{booktabs}

\newif\iftwocol
\twocolfalse 

\newtheorem{theorem}{Theorem}
\newtheorem{lemma}{Lemma}
\newtheorem{proposition}{Proposition}

\newtheorem{setting}{Setting}
\newtheorem{corollary}{Corollary}

\newcommand{\R}{\mathbb{R}}
\newcommand{\eps}{\epsilon}
\newcommand{\EE}[1]{\mathbb{E}\left[{#1}\right]}
\newcommand{\EEst}[2]{\mathbb{E}\left[{#1}\  \middle| \ {#2}\right]}

\newcommand{\PP}[1]{\mathbb{P}\left({#1}\right)}
\newcommand{\PPst}[2]{\mathbb{P}\left({#1}\  \middle| \ {#2}\right)}
\newcommand{\Ppst}[3]{\mathbb{P}_{{#1}}\left({#2}\  \middle| \ {#3}\right)}
\newcommand{\Pp}[2]{\mathbb{P}_{{#1}}\left({#2}\right)}

\newcommand{\one}[1]{{\mathbbm{1}}_{{#1}}}
\newcommand{\iidsim}{\stackrel{\textnormal{iid}}{\sim}}

\newcommand{\cC}{\mathcal{C}}
\newcommand{\cX}{\mathcal{X}}
\newcommand{\cY}{\mathcal{Y}}

  \allowdisplaybreaks

\title{Online conformal prediction with decaying step sizes}
\author{Anastasios N. Angelopoulos$^1$, Rina Foygel Barber$^2$, and Stephen Bates$^3$}
\date{$^1$University of California, Berkeley \quad $^2$University of Chicago\\
$^3$Massachusetts Institute of Technology\\
\texttt{angelopoulos@berkeley.edu}, \texttt{rina@uchicago.edu}, \texttt{s\_bates@mit.edu}\\
\hfill \\
\today}

\begin{document}
\maketitle

\begin{abstract}
We introduce a method for online conformal prediction with decaying step sizes.
Like previous methods, ours possesses a retrospective guarantee of coverage for arbitrary sequences.
However, unlike previous methods, we can simultaneously estimate a population quantile when it exists.
Our theory and experiments indicate substantially improved practical properties: in particular, when the distribution is stable, the coverage is close to the desired level \emph{for every time point}, not just on average over the observed sequence.
\end{abstract}

\section{Introduction}

We study the problem of online uncertainty quantification, such as that encountered in time-series forecasting.
Our goal is to produce a \emph{prediction set} at each time, based on all previous information, that contain the true label with a specified coverage probability.
Such prediction sets are useful to the point of being requirements in many sequential problems, including medicine~\citep{robinson1978sequential}, robotics~\citep{lindemann2023safe}, finance~\citep{mykland2003financial}, and epidemiology~\citep{cramer2022evaluation}.
Given this broad utility, it comes as no surprise that prediction sets have been studied for approximately one hundred years (and possibly more; see Section 1.1 of~\citet{tian2022methods}).

Formally, consider a sequence of data points $(X_t, Y_t)\in\cX\times\cY$, for $t=1,2,\dots$.
At each time $t$, we observe $X_t$ and seek to cover $Y_t$ with a set $\cC_t(X_t)$, which depends on a \emph{base model} trained on all past data (as well as the current feature $X_t$).
After predicting, we observe $Y_t$, and the next time-step ensues.
Note that we have not made any assumptions yet about the data points and their dependencies.

This paper introduces a method for constructing the prediction sets $\cC_t$ that has \textbf{simultaneous best-case and worst-case guarantees}---that is, a  ``best of both worlds'' property.
We will describe the method shortly in Section~\ref{subsec:method}.
Broadly speaking, the method can gracefully handle both arbitrary adversarial sequences data points and also independent and identically distributed (I.I.D.) sequences.
In the former case, our method will remain robust, ensuring that the historical fraction of miscovered labels converges to the desired error rate, $\alpha \in (0,1)$.
In the latter case, our method will converge, eventually producing the optimal prediction sets.
We summarize our results below:
\begin{enumerate}
    \item \textbf{Worst-case guarantee (Theorem~\ref{thm:error_bound}):} When the data points are arbitrary, our algorithm achieves
    \begin{equation}\label{eqn:preview_longrun}
        \frac{1}{T}\sum\limits_{t=1}^T \one{Y_t \in \cC_t(X_t)} \in \left( 1-\alpha \pm \frac{C}{T^{1/2-\epsilon}} \right),
    \end{equation}
    for a constant $C$ and any fixed $\epsilon > 0$.
    We call this a \emph{long-run coverage} guarantee.
    \item \textbf{Best-case guarantee (Theorem~\ref{thm:q_converge}):} When the data points are I.I.D., our algorithm achieves
    \begin{equation}\label{eqn:preview_convergence}
        \lim_{T \to \infty}\PP{Y_T \in \cC_T(X_T)} \to 1-\alpha.
    \end{equation}
    We call this a \emph{convergence guarantee}.
\end{enumerate}

Our algorithm is the first to satisfy both guarantees simultaneously. Moreover, the decaying step size yields more stable behavior than prior methods, as we will see in experiments. See Section~\ref{subsec:related-work} for a discussion of the relationship with other methods, such as those of~\citet{gibbs2021adaptive},~\citet{angelopoulos2023conformalpid}, and~\citet{xu2021conformal}.

\subsection{Method and Setup}
\label{subsec:method}

We now describe our prediction set construction.
Borrowing from conformal prediction, consider a bounded conformal score function $s_t : \cX \times \cY \to [0,B]$, at each time $t$. 
This score $s_t = s_t(X_t,Y_t)$ is large when the predictions of the base model disagree greatly with the observed label; an example would be the residual score, $s_t(x,y) = |y-\hat{f}_t(x)|$, for a model $\hat{f}_t : \cX \to \R$ trained online. 
This concept is standard in conformal prediction~\citep{vovk2005algorithmic}, and we refer the reader to~\citet{angelopoulos2023conformal} for a recent overview.
Given this score function, define
\begin{equation}
\label{eq:set-construction}
    \cC_t(x) = \left\{y\in\cY: s_t(x,y)\leq q_t\right\},
\end{equation}
where the threshold $q_t$ is updated with the rule
\begin{equation}
    \label{eq:q_update}
        q_{t+1}= q_t + \eta_t(\one{Y_t \notin \cC_t(X_t)} - \alpha).
\end{equation}
In particular, if we fail to cover $Y_t$ at time $t$, then the threshold increases to make the procedure slightly more conservative at the next time step (and vice versa).

Familiar readers will notice the similarity of the update step~\eqref{eq:q_update} to that of~\citet{gibbs2021adaptive, bhatnagar2023improved, feldman2023achieving, angelopoulos2023conformalpid}, the main difference being that here, $\eta_t$ can change over time---later on we will see that  $\eta_t \propto t^{-1/2 - \epsilon}$, for some small $ \epsilon \in(0,1/2)$, leads to guarantees~\eqref{eqn:preview_longrun} and~\eqref{eqn:preview_convergence} as described above.
We remark also that the update step for $q_t$ can be interpreted as an online (sub)gradient descent algorithm on the quantile loss
$\rho_{1-\alpha}(t) = (1-\alpha) \max\{t,0\} + \alpha\max\{-t,0\}$ ~\citep{koenker1978regression}, i.e.,  we can equivalently write the update step~\eqref{eq:q_update} as
\[        q_{t+1} = q_t - \eta_t \nabla \rho_{1-\alpha}(s_t - q_t).\]

In this work, we will consider two different settings:
\begin{setting}[Adversarial setting]
    \label{setting:adversarial}
    We say that we are in the adversarial setting if we allow $(X_1,Y_1),(X_2,Y_2),\dots$ to be an arbitrary sequence of elements in $\cX \times \cY$, and $s_1,s_2,\dots$ to be an arbitrary sequence of functions from $\cX \times \cY$ to $[0,B]$.
\end{setting}
\begin{setting}[I.I.D.\ setting]
    \label{setting:iid}
    We say that we are in the I.I.D.\ setting if we require that $(X_t, Y_t) \iidsim P$ for some distribution $P$, and require that the choice of the function $s_t:\cX\times\cY\rightarrow[0,B]$ depends only on $\{(X_r,Y_r)\}_{r<t}$, for each $t$ (i.e., the model is trained online).
\end{setting}
\noindent Of course, any result proved for Setting~\ref{setting:adversarial} will hold for Setting~\ref{setting:iid} as well. 
We remark that Setting~\ref{setting:iid} can be relaxed to allow for randomness in the choice of the score functions $s_t$---our results  for the I.I.D.\ setting will hold as long as the function $s_t$ is chosen independently of $\{(X_r,Y_r)\}_{r\geq t}$.

Our method, like all conformal methods, has coverage guarantees that hold for \emph{any} underlying model and data stream. Still, the quality of the output (e.g., the size of the prediction sets) does critically depend on the quality of the underlying model. This general interplay between conformal methods and models is discussed throughout the conformal literature~\citep[e.g.,][]{vovk2005algorithmic, angelopoulos2023conformal}.

\subsection{Related work}
\label{subsec:related-work}

We begin by reviewing the most closely related literature.
Set constructions of the form in~\eqref{eq:set-construction}, which ``invert'' the score function, are commonplace in conformal prediction~\citep{vovk2005algorithmic}, with $q_t$ chosen as a sample quantile of the previous conformal scores.
However, the exchangeability-based arguments of the standard conformal framework cannot give any guarantees in Setting~\ref{setting:adversarial}.
The idea to set $q_t$ via online gradient descent with a \emph{fixed} step size appears first in~\citet{gibbs2021adaptive}, which introduced online conformal prediction in the adversarial setting. The version we present here builds also on the work of~\citet{bhatnagar2023improved},~\citet{feldman2023achieving}, and ~\citet{angelopoulos2023conformalpid}; in particular,~\citet{angelopoulos2023conformalpid} call the update in~\eqref{eq:q_update} the ``quantile tracker''.
These papers all have long-run coverage guarantees in Setting~\ref{setting:adversarial}, but do not have convergence guarantees in Setting~\ref{setting:iid}.

Subsequent work to these has explored time-varying step sizes that respond to distribution shifts, primarily for the purpose of giving other notions of validity, such as regret analyses~\citep{gibbs2022conformal, zaffran2022adaptive, bastani2022practical, noarov2023high, bhatnagar2023improved}.
From an algorithmic perspective, these methods depart significantly from the update in~\eqref{eq:q_update}, generally by incorporating techniques from online learning---such as strongly adaptive online learning~\citep{daniely2015strongly}, adaptive regret~\citep{gradu2023adaptive}, and adaptive aggregation of experts~\citep{cesa2006prediction}.
To summarize, the long-run coverage and regret bounds in these papers apply to substantially different, usually more complicated algorithms than the simple expression we have in~\eqref{eq:q_update}.
We remark that ``best of both worlds'' guarantees appear in the online learning literature~\citep[e.g.,][]{bubeck2012best,koolen2016combining,zimmert2021tsallis,jin2021best,chen2023optimistic,dann2023best}, where the aim is to find a single algorithm whose \emph{regret} is optimal both in a stochastic setting (i.e., data sampled from a distribution) and in an adversarial setting. A crucial difference, however, is that our paper's guarantees are concerned with inference and predictive coverage, rather than with estimation or regret.

Farther afield from our work, there have been several other explorations of conformal prediction in time-series, but these are quite different.
For example, the works of~\citet{barber2022conformal} and~\citet{chernozhukov2018exact} provide conformal-type procedures with coverage guarantees under certain relaxations of exchangeability; both can provide marginal coverage in Setting~\ref{setting:iid}, but cannot give any guarantees in Setting~\ref{setting:adversarial}.
\citet{xu2021conformal, xu2023sequential} study the behavior of conformal methods under classical nonparametric assumptions such as model consistency and distributional smoothness for its validity, and thus cannot give distribution-free guarantees in Settings~\ref{setting:adversarial} or~\ref{setting:iid}.
\citet{lin2022conformal} studies the problem of cross-sectional coverage for multiple exchangeable time-series.
The online conformal prediction setup was also considered early on by~\citet{vovk2002line} for exchangeable sequences.
These works are not directly comparable to ours, the primary point of difference being the adversarial guarantee we can provide in Setting~\ref{setting:adversarial}. 

Finally, traditional solutions to the prediction set problem have historically relied on Bayesian modeling~\citep[e.g.,][]{foreman2017fast} or distributional assumptions such as autoregression, smoothness, or ergodicity~\citep[e.g.,][]{biau2011sequential}.
A parallel literature on calibration exists in the adversarial sequence model~\citep[e.g.,][]{foster1998asymptotic}.
Our work, like that of~\citet{gibbs2021adaptive}, is clearly related to the literatures on both calibration and online convex optimization~\citep{zinkevich2003online}, and we hope these connections will continue to reveal themselves; our work takes online conformal prediction one step closer to online learning by allowing the use of decaying step sizes, which is typical for online gradient descent.

\subsection{Our contribution}
We provide the first analysis of the online conformal prediction update in~\eqref{eq:q_update} with an arbitrary step size.
Our analysis gives strong long-run coverage bounds for appropriately decaying step sizes, even in the adversarial setting (Setting~\ref{setting:adversarial}).
We also give a simultaneous convergence guarantee in the I.I.D.\ setting (Setting~\ref{setting:iid}), showing that the parameter $q_t$ converges to the optimal value $q^*$. 
Importantly, this type of convergence does \emph{not} hold with a fixed step size (the case previously analyzed in the online conformal prediction literature). In fact, we show that with a fixed step size, online conformal prediction returns meaningless prediction sets (i.e., either $\emptyset$ or $\cY$) infinitely often. From the theoretical point of view, therefore, our method is the first to provide this type of ``best-of-both-worlds'' guarantee. 

While these theoretical results show an improvement (relative to the fixed-step-size method) in an I.I.D.\ setting, from the practical perspective we will see that a decaying step size also enables substantially better results and more stable behavior on real time series data, which lies somewhere between the I.I.D.\ and the adversarial regime.

\section{Main results in the adversarial setting}\label{sec:results_adversarial}

We now present our main results for the adversarial setting, Setting~\ref{setting:adversarial}, which establish long-run coverage guarantees with no assumptions on the data or the score functions.

\subsection{Decreasing step sizes}
Our first main result shows that, for a nonincreasing step size sequence, 
the long-run coverage rate
\begin{equation}
    \label{eq:long-run-coverage}
    \frac{1}{T}\sum_{t=1}^T \one{Y_t\in\cC_t(X_t)}
\end{equation}
will converge to the nominal level $1-\alpha$.

\begin{theorem}\label{thm:error_bound}
Let $(X_1,Y_1),(X_2,Y_2),\dots$ be an arbitrary sequence of data points, and let $s_t:\cX\times\cY\rightarrow[0,B]$ be arbitrary functions. 
Let $\eta_t$ be a positive and nonincreasing sequence of step sizes, and fix an initial threshold $q_1\in[0,B]$.

Then online conformal prediction satisfies
\[\left|\frac{1}{T}\sum_{t=1}^T \one{Y_t\in\cC_t(X_t)} 
- (1-\alpha) \right|
\leq \frac{B+\eta_1}{\eta_T T}\]
for all $T\geq 1$.
\end{theorem}

As a special case, if we choose a constant step size $\eta_t\equiv \eta$ then this result is analogous to \citet[Proposition 4.1]{gibbs2021adaptive}. On the other hand, if we choose $\eta_t\propto t^{-a}$ for some $a\in(0,1)$, then the long-run coverage at time $T$ has error bounded as $\mathcal{O}(\frac{1}{T^{1-a}})$.

\subsection{Arbitrary step sizes}\label{sec:result_for_arbitrary_step_sizes}
As discussed above, if the data appears to be coming from the same distribution then a decaying step size can be advantageous, to stabilize the behavior of the prediction sets over time. However, if we detect a sudden distribution shift and start to lose coverage, we might want to increase the step size $\eta_t$ to recover coverage more quickly. To accommodate this, the above theorem can be generalized to an arbitrary step size sequence, as follows.
\begin{theorem}\label{thm:error_bound_general}
Let $(X_1,Y_1),(X_2,Y_2),\dots$ be an arbitrary sequence of data points, and let $s_t:\cX\times\cY\rightarrow[0,B]$ be arbitrary functions. 
Let $\eta_t$ be an arbitrary positive sequence, and fix an initial threshold $q_1\in[0,B]$.

Then online conformal prediction satisfies
\iftwocol
\begin{multline*}\left|\frac{1}{T}\sum_{t=1}^T \one{Y_t\in\cC_t(X_t)} 
- (1-\alpha) \right|\\
\leq \frac{B+\max_{1\leq t\leq T}\eta_t}{T} \cdot \|\Delta_{1:T}\|_1\end{multline*}
\else
\[\left|\frac{1}{T}\sum_{t=1}^T \one{Y_t\in\cC_t(X_t)} 
- (1-\alpha) \right|
\leq \frac{B+\max_{1\leq t\leq T}\eta_t}{T} \cdot \|\Delta_{1:T}\|_1\]
\fi
for all $T\geq 1$,
where the sequence $\Delta$ is defined with values
\[\Delta_1 = \eta_1^{-1}, \textnormal{\ and \ }\Delta_t = \eta_t^{-1} - \eta_{t-1}^{-1}\textnormal{ for all $t\geq 2$}.\]
\end{theorem}
We can see that Theorem~\ref{thm:error_bound} is indeed a special case of this more general result, because in the case of a nonincreasing sequence $\eta_t$, we have $\max_{1\leq t\leq T}\eta_t = \eta_1$, and
\iftwocol
    \begin{multline*}\|\Delta_{1:T}\|_1 = |\eta_1^{-1}| + \sum_{t=2}^T |\eta_t^{-1} - \eta_{t-1}^{-1}|\\ = \eta_1^{-1} + \sum_{t=2}^T (\eta_t^{-1} - \eta_{t-1}^{-1}) = \eta_T^{-1}.\end{multline*}
\else
    \[
        \|\Delta_{1:T}\|_1 = |\eta_1^{-1}| + \sum_{t=2}^T |\eta_t^{-1} - \eta_{t-1}^{-1}|\\ = \eta_1^{-1} + \sum_{t=2}^T (\eta_t^{-1} - \eta_{t-1}^{-1}) = \eta_T^{-1}.
    \]
\fi
But Theorem~\ref{thm:error_bound_general} can be applied much more broadly. For example, we might allow the step size to decay during long stretches of time when the distribution seems stationary, but then reset to a larger step size whenever we believe the distribution may have shifted. In this case, we can obtain an interpretable error bound from the result of Theorem~\ref{thm:error_bound_general} by observing that
$\|\Delta_{1:T}\|_1 \leq \frac{2N_T}{\min_{1\leq t\leq T}\eta_t}$,
where $N_T = \sum_{t=2}^T \one{\eta_t > \eta_{t-1}}$ is the number of times we increase the step size. Thus, as long as the step size does not decay too quickly, and the number of ``resets'' $N_T$ is $\mathrm{o}(T)$, the upper bound of Theorem~\ref{thm:error_bound_general} will still be vanishing.

\section{Results for I.I.D.\ data}

We now turn to studying the setting of I.I.D.\ data, Setting~\ref{setting:iid}, where $(X_1,Y_1),(X_2,Y_2),\dots$ are sampled I.I.D.\ from some distribution $P$ on $\cX\times\cY$. While Theorems~\ref{thm:error_bound} and~\ref{thm:error_bound_general} show that the coverage of the procedure converges in a weak sense, as in~\eqref{eq:long-run-coverage}, for \emph{any} realization of the data (or even with a nonrandom sequence of data points), we would also like to understand whether the procedure might satisfy stronger notions of convergence with ``nice'' data. Will the sequence of prediction intervals converge in a suitable sense? We will see that decaying step size does indeed lead to convergence, whereas a constant step size leads to oscillating behavior.

In order to make our questions precise, we need to introduce one more piece of notation to capture the notion of coverage at a particular time $t$---the ``instantaneous'' coverage. Let
\[\mathsf{Coverage}_t(q) = \Ppst{P}{s_t(X,Y) \leq q}{s_t},\]
where the probability is calculated with respect to a data point $(X,Y)\sim P$ drawn independently of $s_t$.
Then, at time $t$, the prediction set $\cC_t(X_t)$ has coverage level $\mathsf{Coverage}_t(q_t)$, by construction.
We will see in our results below that for an appropriately chosen decaying step size, $\mathsf{Coverage}_t(q_t)$ will concentrate around $1-\alpha$ over time, while if we choose a constant step size, then $\mathsf{Coverage}_t(q_t)$ will be highly variable.

\subsection{Results with a pretrained score function}
\label{subsec:iid}

To begin, we assume that the score function is pretrained, i.e., that $s_1 = s_2 = \dots$ are all equal to some fixed function $s:\cX\times\cY\rightarrow[0,B]$. The reader should interpret this as the case where the underlying model is not updated online (e.g., $s(x,y) = |y-\hat{f}(x)|$ for a pretrained model $\hat{f}$ that is no longer being updated). This simple case is intended only as an illustration of the trends we might see more generally; in Section~\ref{subsec:iid_online_training} below we will study a more realistic setting, where model training is carried out online as the data is collected.

In this setting, since the score function does not vary with $t$, we have $\mathsf{Coverage}_t(\cdot)\equiv \mathsf{Coverage}(\cdot)$ where 
\[\mathsf{Coverage}(q) = \Pp{P}{s(X,Y) \leq q},\]
i.e., instantaneous coverage at time $t$ is $\mathsf{Coverage}(q_t)$.

First, we will see that choosing a \emph{constant} step size leads to undesirable behavior: while coverage will hold on average over time (as recorded in Theorem~\ref{thm:error_bound} and in the earlier work of \citet{gibbs2021adaptive}), there will be high variability in $\mathsf{Coverage}(q_t)$ over time---for instance, we may have $\cC_t(X_t)=\emptyset$ infinitely often.

\begin{proposition}\label{prop:constant_q}
Let $(X_t,Y_t) \iidsim P$ for some distribution $P$. Suppose also that $s_t \equiv s$ for some fixed function $s : \cX \times \cY \to [0,B]$, and that $\eta_t\equiv \eta$ for a positive constant step size $\eta>0$. Assume also that $\alpha$ is a rational number.

Then online conformal prediction satisfies
\iftwocol
    \[\textnormal{$\mathsf{Coverage}(q_t) = 0$  for infinitely many $t$,}\]
    and
    \[\textnormal{$\mathsf{Coverage}(q_t) = 1$  for infinitely many $t$,}\]
\else
    \[\textnormal{$\mathsf{Coverage}(q_t) = 0$  for infinitely many $t$, and $\mathsf{Coverage}(q_t) = 1$  for infinitely many $t$,}\]
\fi
almost surely.
\end{proposition}
\noindent In other words, even in the simplest possible setting of I.I.D.\ data and a fixed model, we cannot expect convergence of the method if we use a constant step size.

On the other hand, if we choose a sequence of step sizes $\eta_t$ that decays at an appropriate rate (such as $\eta_t \propto t^{-1/2 - \epsilon}$, for some $ \epsilon \in(0,1/2)$, as mentioned earlier) then over time, this highly variable behavior can be avoided. Instead, we will typically see coverage converging to $1-\alpha$ for \emph{each} constructed prediction set $\cC_t(X_t)$, i.e., $\mathsf{Coverage}(q_t)\rightarrow 1-\alpha$.  We will need one more assumption: defining $q^*$ as the $(1-\alpha)$-quantile of $s(X,Y)$, 
we assume that $q^*$ is unique:
\begin{equation}\label{eqn:qstar_unique}
\begin{array}{l}\mathsf{Coverage}(q) < 1-\alpha \textnormal{ for all $q<q^*$}, \\ \mathsf{Coverage}(q) > 1-\alpha \textnormal{ for all $q>q^*$}.\end{array}\end{equation}

\begin{theorem}\label{thm:q_converge}
Let $(X_t,Y_t) \iidsim P$ for some distribution $P$. Suppose also that $s_t \equiv s$ for some fixed function $s : \cX \times \cY \to [0,B]$. Assume that $\eta_t$ is a fixed nonnegative step size sequence satisfying
\begin{equation}\label{eqn:eta_t_sums}\sum_{t=1}^\infty \eta_t = \infty, \quad \sum_{t=1}^\infty \eta_t^2 < \infty.\end{equation}
Assume also that $q^*$ is unique as in~\eqref{eqn:qstar_unique}.

Then online conformal prediction satisfies
\[q_t  \rightarrow q^*\textnormal{ almost surely}.\]
\end{theorem}
With an additional assumption, this immediately implies convergence of the coverage, $\mathsf{Coverage}(q_t)$:
\begin{corollary}\label{cor:cor1}
    Under the setting and assumptions of Theorem~\ref{thm:q_converge}, assume also that $s(X,Y)$ has a continuous distribution (under $(X,Y)\sim P$). Then 
    \[\mathsf{Coverage}(q_t)\rightarrow 1-\alpha\textnormal{ almost surely}.\]
\end{corollary}
That is, instead of the high variance in coverage incurred by a constant step size (as in Proposition~\ref{prop:constant_q}), here the coverage converges to the nominal level $1-\alpha$. Finally, with  additional assumptions, we can also characterize the \emph{rate} at which the threshold $q_t$ converges to $q^*$:
\begin{proposition}\label{prop:converge_rate}
    Under the setting and assumptions of Corollary~\ref{cor:cor1}, assume also that the distribution of $s(X,Y)$ (under $(X,Y)\sim P$) has density lower-bounded by $\gamma$ in the range $[q^*-\delta,q^*+\delta]$, for some $\gamma,\delta>0$. Take the step size sequence $\eta_t = ct^{-1/2-\eps}$, for some $c>0$ and $\eps\in(0,1/2)$. Then it holds for all $t\geq 1$ that
    \[\EE{(q_t-q^*)^2}\leq bt^{-1/2-\eps},\]
    where $b$ is a constant that depends only on $B,\gamma,\delta,c,\eps$.
\end{proposition}

\subsection{Results with online training of the score function}
\label{subsec:iid_online_training}

The result of Theorem~\ref{thm:q_converge} above is restricted to a very simple setting, where the score functions are given by $s_t\equiv s$ for some fixed $s$, i.e., we are using a pretrained model. We now consider the more interesting setting where the model is trained online.
Formally, we consider Setting~\ref{setting:iid} where we allow the score function $s_t$ to depend arbitrarily on the data observed \emph{before time $t$}, i.e., on $\{(X_r,Y_r)\}_{r<t}$. 

First, we will consider a constant step size $\eta_t\equiv \eta$.
\begin{proposition}\label{prop:constant_q_online_training}
Let $(X_t,Y_t) \iidsim P$ for some distribution $P$, and assume the score functions $s_t:\cX\times\cY\rightarrow[0,B]$ are trained online. Let $\eta_t\equiv \eta$ for a positive constant step size $\eta>0$.

Then online conformal prediction satisfies
\[\lim\inf_{t\rightarrow\infty}\mathsf{Coverage}_t(q_t) =0, \quad \lim\sup_{t\rightarrow\infty}\mathsf{Coverage}_t(q_t) =1\]
almost surely.
\end{proposition}
This result is analogous to Proposition~\ref{prop:constant_q} for the case of a pretrained score function (but with a slightly weaker conclusion due to the more general setting). As before, the conclusion we draw is that a constant step size inevitably leads to high variability in $\mathsf{Coverage}_t(q_t)$.

On the other hand, if we take a decaying step size,  Theorem~\ref{thm:q_converge} established a convergence result given a pretrained score function. We will now see that similar results hold in for the online setting as long as the model converges in some sense. In many settings, we might expect $s_t$ to converge to some score function $s$---for example, if our fitted regression functions, $\hat{f}_t$, converge to some ``true'' model $f^*$, then $s_t(x,y) = |y-\hat{f}_t(x)|$ converges to $s(x,y) = |y-f^*(x)|$. As before, we let $\mathsf{Coverage}(q) = \Pp{P}{s(X,Y)\leq q}$, and write $q^*$ to denote the $(1-\alpha)$-quantile of this distribution.
We now extend the convergence results of Theorem~\ref{thm:q_converge} to this setting. 

\begin{theorem}\label{thm:q_converge2}
Let $(X_t,Y_t) \iidsim P$ for some distribution $P$, and assume the score functions $s_t$ are trained online. Assume that $\eta_t$ is a fixed nonnegative step size sequence satisfying~\eqref{eqn:eta_t_sums}.
Let $s:\cX\times\cY\rightarrow[0,B]$ be a fixed score function, and assume that $q^*$ is unique as in~\eqref{eqn:qstar_unique}.

Then online conformal prediction satisfies the following statement almost surely:\footnote{We use $s_t\stackrel{\textnormal{d}}{\rightarrow} s$ in the sense of convergence in distribution under $(X,Y)\sim P$, while treating the $s_t$'s as fixed. Specifically, we are assuming $\mathsf{Coverage}_t(q)\rightarrow \mathsf{Coverage}(q)$, for all $q\in\mathbb{R}$ at which $\mathsf{Coverage}(q)$ is continuous.}
\[\textnormal{If $s_t\stackrel{\textnormal{d}}{\rightarrow} s$, then $q_t  \rightarrow q^*$.}\]
\end{theorem}
As in the previous section, an additional assumption implies convergence of the coverage, $\mathsf{Coverage}_t(q_t)$:
\begin{corollary}\label{cor:cor2}
    Under the setting and assumptions of Theorem~\ref{thm:q_converge2}, assume also that $s(X,Y)$ has a continuous distribution (under $(X,Y)\sim P$).  Then online conformal prediction satisfies the following statement almost surely: 
    \[\textnormal{If $s_t\stackrel{\textnormal{d}}{\rightarrow} s$, then $\mathsf{Coverage}_t(q_t)  \rightarrow 1-\alpha$.}\]
\end{corollary}

To summarize, the results of this section show that the coverage of each prediction set $\cC_t(X_t)$, given by $\mathsf{Coverage}_t(q_t)$, will converge even in a setting where the model is being updated in a streaming fashion, as long as the fitted model itself converges over time.

In particular, if we choose $\eta_t\propto t^{-1/2-\eps}$ for some $\eps\in(0,1/2)$, then in the adversarial setting the long-run coverage error is bounded as $\mathcal{O}(\frac{1}{T^{1/2-\eps}})$ by Theorem~\ref{thm:error_bound}, while in the I.I.D.\ setting, Theorem~\ref{thm:q_converge2} guarantees convergence. In other words, this choice of $\eta_t$ simultaneusly achieves both types of guarantees.

While the results of this section have assumed I.I.D.\ data, the proof techniques used here can be extended to handle broader settings---for example, a stationary time series, where despite dependence we may still expect to see convergence over time. We leave these extensions to future work.

\section{Experiments}

We include two experiments: an experiment on the Elec2 dataset~\citep{harries1999splice} where the data shows significant distribution shift over time, and an experiment on Imagenet~\citep{deng2009imagenet} where the data points are exchangeable.\footnote{Code to reproduce these experiments is available at \url{https://github.com/aangelopoulos/online-conformal-decaying}.}

The experiments are run with two different choices of step size for online conformal: first,
a fixed step size ($\eta_t \equiv 0.05$); and second, a decaying step size ($\eta_t = t^{-1/2 - \epsilon}$ with $\epsilon=0.1$). 
We also compare to an oracle method, where online conformal is run with $q^*$ in place of $q_t$ at each time $t$, and $q^*$ is chosen to be the value that gives $1-\alpha$ average coverage over the entire sequence $t=1,\dots,T$.
All methods are run with $\alpha=0.1$.

\subsection{Results}
\begin{figure*}[t]
    \centering
    \includegraphics[width=\textwidth]{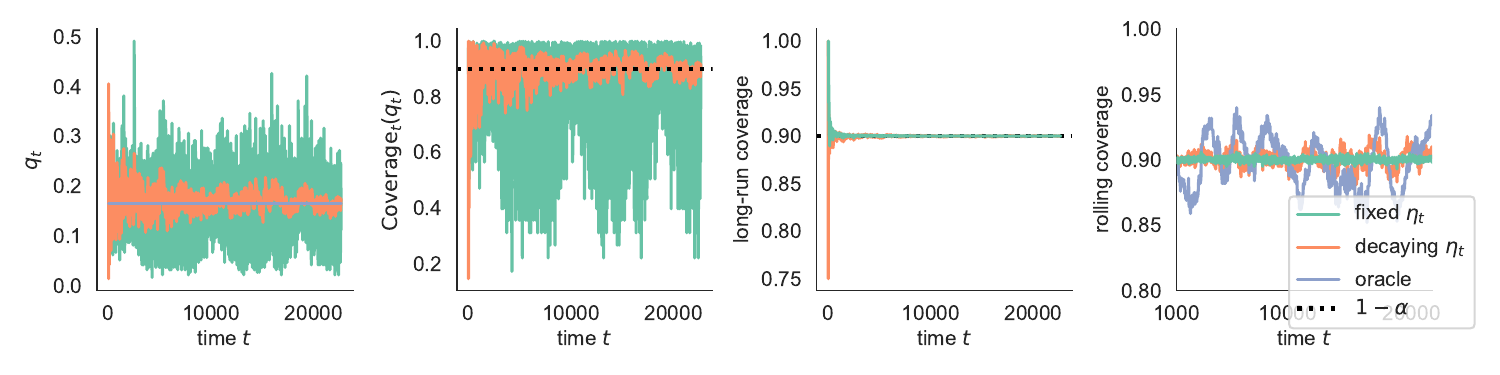}\vspace{-.2in}
    \caption{\textbf{Elec2 results.} From left to right, the panels display the following (over all times $t$): first, the value of the threshold $q_t$; second, the instantaneous coverage $\mathsf{Coverage}_t(q_t)$; third, the long-run coverage $\frac{1}{t}\sum_{r=1}^t \one{Y_r\in\cC_r(X_r)}$; and fourth, the rolling coverage, averaged over a rolling window of 1000 time points.  
    }
    \label{fig:elec2-results}
    \includegraphics[width=\textwidth]{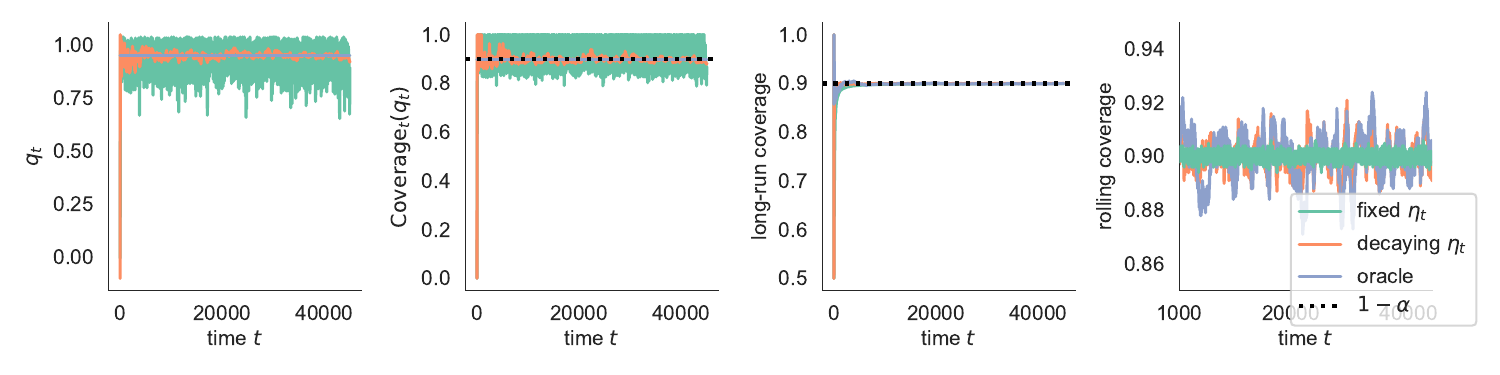}\vspace{-.2in}
    \caption{\textbf{Imagenet results.} Same details as for Figure~\ref{fig:elec2-results}.
    }
    \label{fig:imagenet-results}
\end{figure*}

Figures~\ref{fig:elec2-results} and~\ref{fig:imagenet-results}  display the results of the experiment for the Elec2 data and the Imagenet data, respectively. We now discuss our findings.

\paragraph{The thresholds $q_t$.}
The first panel of each figure plots the value of the threshold $q_t$ over time $t$. 
We can see that the procedure with a fixed step size has significantly larger fluctuations in the quantile value as compared to the decaying step size procedure.

\paragraph{The instantaneous coverage $\mathsf{Coverage}_t(q_t)$.}
The second panel of each figure plots the value of the instantaneous coverage $\mathsf{Coverage}_t(q_t)$ over time $t$. 
For each dataset, since the true data distribution is unknown, we estimate $\mathsf{Coverage}_t(q_t)$ using a holdout set. We observe that $\mathsf{Coverage}_t(q_t)$ is substantially more stable for the decaying step size as compared to  fixed step size in both experiments.
While $\mathsf{Coverage}_t(q_t)$ concentrates closely around the nominal level $1-\alpha$ for decaying $\eta_t$, for fixed $\eta_t$ the coverage level oscillates and does not converge.

\paragraph{Long-run coverage and rolling coverage.}
The third panel of each figure plots the value of the long-run coverage, $\frac{1}{r}\sum_{r=1}^t\one{Y_r\in\cC_r(X_r)}$, over time $t$.
We see that the long-run coverage converges quickly to $1-\alpha$ for all methods, and we cannot differentiate between them in this plot. 

Consequently, in the fourth panel of each figure, we also plot the ``rolling'' coverage, which computes coverage rate averaged over a rolling window of 1000 time points. We can see that this measure is tighter around $1-\alpha$ for the fixed step size procedure; for the decaying step size procedure, rolling coverage fluctuates more, but is not larger than the fluctuations for the oracle method.
At first glance, it might appear that having lower variance in the rolling coverage indicates that the fixed step size procedure is actually performing \emph{better} than decaying step size---but this is not the case. The low variance with fixed $\eta_t\equiv \eta$ is due to \emph{overcorrecting}. For example, if we have several miscoverage events in a row (which can happen by random chance, even with the oracle intervals), then the fixed-step-size method will necessarily return an overly wide interval (e.g., $\cC_t(X_t)=\R$) to give certain coverage at the next time step. 
Thus, the fixed-step-size method ensures low variance in rolling coverage at the cost of extremely high variance in the width and instantaneous coverage of the interval $\cC_t(X_t)$
 at each time 
$t$. 
This type of overcorrection is undesirable.

\subsection{Implementation details for Elec2 Time Series}
The Elec2~\citep{harries1999splice} dataset is a time-series of 45312 hourly measurements of electricity demand in New South Wales, Australia. We use even-numbered time points as the time series, and odd-numbered time points as a holdout set for estimating $\mathsf{Coverage}_t(q_t)$.
The demand measurements are normalized to lie in the interval $Y_t \in [0,1]$.
The covariate vector $X_t = (Y_1, \ldots, Y_{t-1})$ is the sequence of all previous demand values. 
The forecast $\hat{Y}_t$ is one-day-delayed moving average of $Y_t$ (i.e., at time $t$, our predicted value $\hat{Y}_t$ is given by the average of observations taken between 24 and 48 hours earlier), and the score is $s_t(X_t,Y_t) = |Y_t - \hat{Y}_t|$.

\subsection{Implementation details for Imagenet}
The Imagenet~\citep{deng2009imagenet} is a standard computer vision dataset of natural images.
We take the 50000 validation images of Imagenet 2012 and treat them as a time series for the purpose of evaluating our methods.
Because the validation split of Imagenet is shuffled, this comprises an exchangeable time series.
We use 45000 points for the time series, and the remaining 5000 points as a holdout set for estimating $\mathsf{Coverage}_t(q_t)$.
As the score function, we use $s_t(X_t,Y_t) = 1-\max_{y \in \cY}\hat{f}(X_t)_y$ (here $\max_{y \in \cY}\hat{f}(X_t)_y$ is the softmax score of the pretrained ResNet-152 model).

\subsection{Additional experiments}
As discussed in Section~\ref{sec:result_for_arbitrary_step_sizes}, in applications where the distribution of the data may drift or may have changepoints, it might be beneficial to allow $\eta_t$ to increase at times to allow for updates in the learning process. To study this empirically, in the Appendix, we include additional experiments in a broader range of settings---we test over 3000 real datasets, and compare the fixed step size method, the decaying step size method, and a ``decay+adapt'' version of our method where the sequence $\eta_t$ adapts to trends in the data (decaying if the distribution of the data appears stationary, but increasing if distribution shift is detected).

\section{Proofs}
In this section, we prove Theorems~\ref{thm:error_bound},~\ref{thm:error_bound_general},~\ref{thm:q_converge}, and~\ref{thm:q_converge2}, and Propositions~\ref{prop:constant_q} and~\ref{prop:constant_q_online_training}. All other results are proved in the Appendix. 
\subsection{Proof of Theorems~\ref{thm:error_bound} and~\ref{thm:error_bound_general}}

First, we need a lemma to verify that the values $q_t$ are uniformly bounded over all $t$. This result is essentially the same as that in Lemma 4.1 of \citet{gibbs2021adaptive}, except extended to the setting of decaying, rather than constant, step size. The proof is given in the Appendix.
\begin{lemma}\label{lem:qt_bound}
Let $(X_1,Y_1),(X_2,Y_2),\dots$ be an arbitrary sequence of data points, and let $s_t:\cX\times\cY\rightarrow[0,B]$ be arbitrary functions. 
Let $\eta_t$ be an arbitrary nonnegative sequence, and fix an initial threshold $q_1\in[0,B]$.

Then online conformal prediction satisfies
\begin{equation}\label{eqn:qt_bounded_range}-\alpha M_{t-1}\leq q_t \leq B + (1-\alpha) M_{t-1} \textnormal{ for all $t\geq 1$},\end{equation}
where $M_0=0$, and $M_t = \max_{1\leq r\leq t} \eta_r$ for each $t\geq 1$.
\end{lemma}

We are now ready to prove the theorems. As discussed earlier, Theorem~\ref{thm:error_bound} is simply a special case, so we only prove the more general result Theorem~\ref{thm:error_bound_general}.

By definition of $\Delta$, we have
$\eta_t^{-1} = \sum_{r=1}^t \Delta_r$
for all $t\geq 1$. 
We then calculate
\begin{align*}
    &\left|\frac{1}{T}\sum_{t=1}^T \one{Y_t\in\cC_t(X_t)} 
- (1-\alpha) \right|
 =\left|\frac{1}{T}\sum_{t=1}^T \one{Y_t\not\in\cC_t(X_t)} 
- \alpha \right|\\
&=\left|\frac{1}{T}\sum_{t=1}^T \left(\sum_{r=1}^t \Delta_r\right) \cdot \eta_t \left(\one{Y_t\not\in\cC_t(X_t)} 
- \alpha\right) \right|\\
&=\left|\frac{1}{T}\sum_{r=1}^T \Delta_r \left(\sum_{t=r}^T \eta_t \left(\one{Y_t\not\in\cC_t(X_t)} 
- \alpha\right) \right)\right|\\
&=\left|\frac{1}{T}\sum_{r=1}^T \Delta_r \left(q_{T+1} - q_r\right)\right|\textnormal{ by~\eqref{eq:q_update}}\\
&\leq \frac{1}{T}\sum_{r=1}^T |\Delta_r| \cdot\max_{1\leq r\leq T} \left|q_{T+1} - q_r\right|\\
&\leq \frac{1}{T} \cdot \|\Delta_{1:T}\|_1 \cdot (B + \max_{1\leq t\leq T}\eta_t),
\end{align*}
where the last step holds since $q_r,q_{T+1}$ are bounded by Lemma~\ref{lem:qt_bound}.

\subsection{Proof of Proposition~\ref{prop:constant_q_online_training}}
First we prove that $\lim\sup_{t\rightarrow\infty} \mathsf{Coverage}_t(q_t) = 1$ almost surely. Equivalently, for any fixed $\eps>0$, we need to prove that $\PP{\lim\sup_{t\rightarrow\infty} \mathsf{Coverage}_t(q_t)< 1-\eps} =0$.

We begin by constructing a useful coupling between the online conformal process, and a sequence of I.I.D.\ uniform random variables. For each $t\geq 1$, define
\[U_t \sim \begin{cases} \mathsf{Uniform}[0,\mathsf{Coverage}_t(q_t)], & \textnormal{ if }Y_t \in\cC_t(X_t),\\
\mathsf{Uniform}[\mathsf{Coverage}_t(q_t),1], & \textnormal{ if }Y_t \not\in\cC_t(X_t),\end{cases}
\]
drawn independently for each $t$ after conditioning on all the data, $\{(X_t,Y_t)\}_{t\geq 1}$. 
Since $\PPst{Y_t \in\cC_t(X_t)}{\{(X_r,Y_r)\}_{r<t}} = \mathsf{Coverage}_t(q_t)$ by construction, we can verify that $U_t\iidsim \mathsf{Uniform}[0,1]$.

Next fix any integer 
$N\geq\frac{B+\eta\alpha}{\eta(1-\alpha)}$.
Let $A_i$ be the event that 
\[U_t > 1-\eps \textnormal{ for all $(i-1)N<t\leq iN$}.\]
Since the $U_t$'s are I.I.D.\ uniform random variables, we have $\PP{A_i} = \eps^N$ for each $i$, and the events $A_i$ are mutually independent. Therefore, by the second Borel--Cantelli lemma,
$\PP{\sum_{i\geq 1}\one{A_i} = \infty}=1$.
Now we claim that
\begin{equation}\label{eq:from_Ai_to_coverage}\textnormal{If $A_i$ occurs then }\max_{(i-1)N<t\leq iN+1}\mathsf{Coverage}_t(q_t) > 1-\eps.\end{equation}
Suppose that $A_i$ holds and that  $\mathsf{Coverage}_t(q_t)\leq 1-\eps$ for all $t$ in the range $(i-1)N<t\leq iN$. Then by construction of the $U_t$'s, we have $Y_t\not\in\cC_t(X_t)$ for all $(i-1)N<t\leq iN$. Therefore by~\eqref{eq:q_update},
\iftwocol
    \begin{multline*}q_{iN+1} = q_{(i-1)N+1} + \sum_{t=(i-1)N+1}^{iN} \eta(\one{Y_t\not\in\cC_t(X_t)} - \alpha) \\= q_{(i-1)N+1} + N\cdot \eta(1-\alpha) \geq B,\end{multline*}
\else
    \[q_{iN+1} = q_{(i-1)N+1} + \sum_{t=(i-1)N+1}^{iN} \eta(\one{Y_t\not\in\cC_t(X_t)} - \alpha) = q_{(i-1)N+1} + N\cdot \eta(1-\alpha) \geq B,\]
\fi
where the last step holds by our choice of $N$, together with the fact that $q_{(i-1)N+1}\geq -\alpha\eta$ by Lemma~\ref{lem:qt_bound}. But since the score function $s_{iN+1}$ takes values in $[0,B]$ by assumption, we therefore have $\mathsf{Coverage}_{iN+1}(q_{iN+1}) \geq \mathsf{Coverage}_{iN+1}(B) = 1$.
Therefore, we have verified the claim~\eqref{eq:from_Ai_to_coverage}.

Since $A_i$ occurs for infinitely many $i$, almost surely, by~\eqref{eq:from_Ai_to_coverage} we therefore have $\lim\sup_{t\rightarrow\infty} \mathsf{Coverage}_t(q_t)\geq 1-\eps$, almost surely, as desired. Since $\eps>0$ is arbitrary, this completes the proof that $\lim\sup_{t\rightarrow\infty} \mathsf{Coverage}_t(q_t) = 1$ almost surely.

Finally, a similar argument verifies  $\lim\inf_{t\rightarrow\infty} \mathsf{Coverage}(q_t) = 0$ almost surely.

\subsection{Proof of Proposition~\ref{prop:constant_q}}
Since $s_t\equiv s$, we have $\mathsf{Coverage}_t(q_t)=\mathsf{Coverage}(q_t)$, for each $t$.
By Proposition~\ref{prop:constant_q_online_training}, $\lim\inf_{t\rightarrow\infty}\mathsf{Coverage}(q_t) = 0$ and $\lim\sup_{t\rightarrow\infty}\mathsf{Coverage}(q_t) = 1$, almost surely.
Since we have assumed that $\alpha$ is a rational number, by the definition of the procedure~\eqref{eq:q_update}, all values $q_t$ must lie on a discrete grid (i.e., if $\alpha=k/K$ for some integers $k,K$ then, for all $t$, $q_t - q_1$ must be an integer multiple of $\eta/K$). Moreover, by Lemma~\ref{lem:qt_bound}, $q_t$ is uniformly bounded above and below for all $t$, so $q_t$ can only take finitely many values. This implies $\mathsf{Coverage}(q_t)$ also can take only finitely many values, and in particular, this means that if $\lim\inf_{t\rightarrow\infty}\mathsf{Coverage}(q_t) = 0$ (respectively, if $\lim\sup_{t\rightarrow\infty} \mathsf{Coverage}(q_t) = 1$) then $\mathsf{Coverage}(q_t)=0$ (respectively, $\mathsf{Coverage}(q_t)=1$) for infinitely many $t$.

\subsection{Proofs of Theorems~\ref{thm:q_converge} and~\ref{thm:q_converge2}}
We observe that 
Theorem~\ref{thm:q_converge} is simply a special case of Theorem~\ref{thm:q_converge2} (obtained by taking $s_t\equiv s$ for all $t$), so we only need to prove Theorem~\ref{thm:q_converge2}.

First, consider the sequence
\[Z_t=\sum_{r=1}^t \eta_r (\one{Y_r\in\cC_r(X_r)} - \mathsf{Coverage}_r(q_r)).\]
Define events $\mathcal{E}_Z$, the event that $\lim_{t\rightarrow\infty} Z_t$ exists, and $\mathcal{E}_s$, the event that $s_t\stackrel{\textnormal{d}}{\rightarrow} s$. 
In the Appendix, we will verify that
\begin{equation}\label{eqn:martingale}
\textnormal{$\lim_{t\rightarrow\infty} Z_t$ exists, almost surely},
\end{equation}
i.e., $\PP{\mathcal{E}_Z}=1$, using martingale theory. 

To establish the theorem, then, it suffices for us to verify that on the event $\mathcal{E}_Z\cap\mathcal{E}_s$, it holds that $q_t\rightarrow q^*$. 
From this point on, we assume that $\mathcal{E}_Z$ and $\mathcal{E}_s$ both hold.

Fix any $\eps>0$. Since $q\mapsto \mathsf{Coverage}(q)$ is monotone, it can have at most countably infinitely many discontinuities. Without loss of generality, then, we can assume that this map is continuous at $q=q^*-\eps/3$ and at $q=q^*+\eps/3$ (by taking a smaller value of $\eps$ if needed).

First, since $Z_t$ converges, we can find some finite time $T_1$ such that
\iftwocol
    \begin{multline}\label{eqn:sum_tail}\sup_{t'\geq t\geq T_1}\left|\sum_{r=t}^{t'} \eta_r (\one{Y_r\in\cC_r(X_r)} - \mathsf{Coverage}_r(q_r))\right| \\ =\sup_{t'\geq t\geq T_1} \left|Z_{t'} - Z_{t-1}\right| \leq \frac{\eps}{3}.\end{multline}
\else
    \begin{equation}\label{eqn:sum_tail}\sup_{t'\geq t\geq T_1}\left|\sum_{r=t}^{t'} \eta_r (\one{Y_r\in\cC_r(X_r)} - \mathsf{Coverage}_r(q_r))\right| \\ =\sup_{t'\geq t\geq T_1} \left|Z_{t'} - Z_{t-1}\right| \leq \frac{\eps}{3}.\end{equation}
\fi
Moreover, since $\sum_t\eta_t^2<\infty$, we have $\eta_t\rightarrow0$ and so we can find some finite time $T_2$ such that $\eta_t \leq \frac{\eps}{3}$ for all $t\geq T_2$. Furthermore, on $\mathcal{E}_s$, we have
$\mathsf{Coverage}_t(q) \rightarrow \mathsf{Coverage}(q)$, at each $q=q^*\pm \eps/3$. Thus we can find some finite time $T_3$ and some $\delta>0$ such that
\begin{equation}\label{eqn:coverage_eps_delta}\mathsf{Coverage}_t(q^*-\eps/3) \leq 1 - \alpha - \delta\end{equation} for all $t\geq T_3$ (we are using the fact that $\mathsf{Coverage}(q^*-\eps/3) < 1-\alpha$ by~\eqref{eqn:qstar_unique}).
Similarly we can find a finite $T_4$ and some $\delta'>0$ such that $\mathsf{Coverage}_t(q^*+\eps/3) \geq 1 - \alpha + \delta'$ for all $t\geq T_4$.
Let $T= \max\{T_1,T_2,T_3,T_4\}$.

We will now split into cases. If it does not hold that $q_t\in q^*\pm \eps$ for all sufficiently large $t$, then one of the following cases must hold:
\begin{itemize}
    \item \textbf{Case 1a:} $q_t < q^* - \eps/3$ for all $t\geq T$.
    \item \textbf{Case 1b:} $q_t > q^* + \eps/3$ for all $t\geq T$.
    \item \textbf{Case 2a:} for some $t' \geq t \geq T$, it holds that $q_t \geq q^* - \eps/3$ and $q_{t'} < q^* - \eps$.
    \item \textbf{Case 2b:} for some $t' \geq t \geq T$, it holds that $q_t \leq q^* + \eps/3$ and $q_{t'} > q^* + \eps$.
\end{itemize}

\noindent We now verify that each case is impossible.

\paragraph{Case 1a is impossible.} We have
\begin{align*}
   & q^* - \frac{\eps}{3} - q_T
    \geq \sup_{t> T} q_t - q_T\\
    &= \sup_{t> T} \sum_{r=T}^{t-1} \eta_r (\one{Y_r\not\in\cC_r(X_r)} - \alpha)\textnormal{ \ by~\eqref{eq:q_update}}\\
    &\geq \sup_{t> T} \sum_{r=T}^{t-1} \eta_r \left((1-\alpha) - \mathsf{Coverage}_r(q_r)\right)- \frac{\eps}{3}\textnormal{ \ by~\eqref{eqn:sum_tail}}\\
    &\geq \sup_{t> T} \left\{\left[\sum_{r=T}^{t-1} \eta_r \cdot \delta \right]- \frac{\eps}{3}\right\},
\end{align*} 
where the last step holds since $q_r < q^*-\eps/3$ for $r\geq T$, and $q\mapsto \mathsf{Coverage}_r(q)$ is nondecreasing, and so we have
\begin{equation}\label{eqn:coverage_eps_delta_2}\mathsf{Coverage}_r(q_r) \leq \mathsf{Coverage}_r(q^*-\eps/3) \leq 1-\alpha-\delta,\end{equation}
by~\eqref{eqn:coverage_eps_delta}.
Since $\sum_r \eta_r = \infty$, we therefore have that $q^* - \frac{\epsilon}{3} - q_T \geq \infty$, which is a contradiction.

\paragraph{Case 1b is impossible.} This proof is analogous to the proof for Case 1a.

\paragraph{Case 2a is impossible.} First, by assumption for this case, we can find a unique time $t''\geq T$ such that 
\[\begin{cases} q_{t''} \geq q^* -\eps/3, \\
q_r < q^* - \eps/3\textnormal{ for all $t'' < r < t'$},\\
q_{t'} < q^* - \eps.\end{cases}\]
In other words, $t''$ is the last time before time $t'$ when the threshold is $\geq q^*-\eps/3$.
Then we have 
\begin{align*}
    &-\frac{2\eps}{3}
     > q_{t'} - q_{t''}
    =  \sum_{r=t''}^{t'-1} \eta_r \left(\one{Y_r\not\in\cC_r(X_r)} - \alpha\right)\textnormal{ \ by~\eqref{eq:q_update}}\\
    &\geq \left[  \sum_{r=t''}^{t'-1} \eta_r \left((1-\alpha) - \mathsf{Coverage}_r(q_r)\right)\right] - \frac{\eps}{3}\textnormal{ \ by~\eqref{eqn:sum_tail}}\\
    &\geq -\eta_{t''} +  \left[  \sum_{r=t''+1}^{t'-1} \eta_r \left((1-\alpha) - \mathsf{Coverage}_r(q_r)\right)\right] - \frac{\eps}{3}\\
    &\geq - \eta_{t''} - \frac{\eps}{3}\textnormal{ by~\eqref{eqn:coverage_eps_delta_2}}.
\end{align*}
But since $\eta_{t''} \leq \eps/3$ (because $t''\geq T$), we have therefore reached a contradiction.

\paragraph{Case 2b is impossible.} This proof is analogous to the proof for Case 2a.

We have verified that all four cases are impossible. Therefore, $q_t\in q^*\pm \eps$ for all sufficiently large $t$. Since $\eps>0$ is arbitrarily small, this completes the proof.

\section{Discussion}

Our paper analyzes online conformal prediction that with a decaying step size, enabling simultaneous guarantees of convergence for I.I.D.\ sequences and long-run coverage for adversarial ones.
Moreover, it helps further unify online conformal prediction with online learning and online convex optimization, since decaying step sizes are known to have desirable properties and hence standard for the latter. 
Of course, the usefulness of the method will rely on choosing score functions that are well suited to the (possibly time-varying) data distribution, and choosing step sizes that decay at an appropriate rate and perhaps adapt to the level of distribution shift---building a better understanding of how to make these choices in practice is crucial for achieving informative and stable prediction intervals. Many additional open questions about extending the methodology to broader settings and understanding connections to other tools remain. 
In particular, we expect fruitful avenues of future inquiry would be: (1) to extend this analysis to online risk control, as in~\citet{feldman2021improving}; (2) to adapt our analysis of Theorem~\ref{thm:q_converge} to deal with stationary or slowly moving time-series which may not be I.I.D.\ but are slowly varying enough to permit estimation; and (3) to further understand the connection between this family of techniques and the theory of online learning.
\subsection*{Acknowledgements}
The authors thank Isaac Gibbs, Ryan Tibshirani, and Margaux Zaffran for helpful feedback on this work. R.F.B.\ was partially supported by the National Science Foundation via grant DMS-2023109, and by the Office of Naval Research via grant N00014-20-1-2337.

\bibliographystyle{plainnat}
\bibliography{bib}

\begin{thebibliography}{38}
\providecommand{\natexlab}[1]{#1}
\providecommand{\url}[1]{\texttt{#1}}
\expandafter\ifx\csname urlstyle\endcsname\relax
  \providecommand{\doi}[1]{doi: #1}\else
  \providecommand{\doi}{doi: \begingroup \urlstyle{rm}\Url}\fi

\bibitem[Angelopoulos and Bates(2023)]{angelopoulos2023conformal}
Anastasios~N Angelopoulos and Stephen Bates.
\newblock Conformal prediction: A gentle introduction.
\newblock \emph{Foundations and Trends{\textregistered} in Machine Learning}, 16\penalty0 (4):\penalty0 494--591, 2023.

\bibitem[Angelopoulos et~al.(2023)Angelopoulos, Candes, and Tibshirani]{angelopoulos2023conformalpid}
Anastasios~N. Angelopoulos, Emmanuel Candes, and Ryan Tibshirani.
\newblock Conformal {PID} control for time series prediction.
\newblock In \emph{Neural Information Processing Systems}, 2023.

\bibitem[Barber et~al.(2022)Barber, Candes, Ramdas, and Tibshirani]{barber2022conformal}
Rina~Foygel Barber, Emmanuel~J Candes, Aaditya Ramdas, and Ryan~J Tibshirani.
\newblock Conformal prediction beyond exchangeability.
\newblock \emph{arXiv:2202.13415}, 2022.

\bibitem[Bastani et~al.(2022)Bastani, Gupta, Jung, Noarov, Ramalingam, and Roth]{bastani2022practical}
Osbert Bastani, Varun Gupta, Christopher Jung, Georgy Noarov, Ramya Ramalingam, and Aaron Roth.
\newblock Practical adversarial multivalid conformal prediction.
\newblock \emph{Advances in Neural Information Processing Systems}, 35:\penalty0 29362--29373, 2022.

\bibitem[Bhatnagar et~al.(2023)Bhatnagar, Wang, Xiong, and Bai]{bhatnagar2023improved}
Aadyot Bhatnagar, Huan Wang, Caiming Xiong, and Yu~Bai.
\newblock Improved online conformal prediction via strongly adaptive online learning.
\newblock \emph{arXiv preprint arXiv:2302.07869}, 2023.

\bibitem[Biau and Patra(2011)]{biau2011sequential}
G{\'e}rard Biau and Beno{\^\i}t Patra.
\newblock Sequential quantile prediction of time series.
\newblock \emph{IEEE Transactions on Information Theory}, 57\penalty0 (3):\penalty0 1664--1674, 2011.

\bibitem[Bubeck and Slivkins(2012)]{bubeck2012best}
S{\'e}bastien Bubeck and Aleksandrs Slivkins.
\newblock The best of both worlds: Stochastic and adversarial bandits.
\newblock In \emph{Conference on Learning Theory}, pages 42--1. JMLR Workshop and Conference Proceedings, 2012.

\bibitem[Cesa-Bianchi and Lugosi(2006)]{cesa2006prediction}
Nicolo Cesa-Bianchi and G{\'a}bor Lugosi.
\newblock \emph{Prediction, learning, and games}.
\newblock Cambridge university press, 2006.

\bibitem[Chen et~al.(2023)Chen, Tu, Zhao, and Zhang]{chen2023optimistic}
Sijia Chen, Wei-Wei Tu, Peng Zhao, and Lijun Zhang.
\newblock Optimistic online mirror descent for bridging stochastic and adversarial online convex optimization.
\newblock In \emph{International Conference on Machine Learning}, pages 5002--5035. PMLR, 2023.

\bibitem[Chernozhukov et~al.(2018)Chernozhukov, W{\"u}thrich, and Yinchu]{chernozhukov2018exact}
Victor Chernozhukov, Kaspar W{\"u}thrich, and Zhu Yinchu.
\newblock Exact and robust conformal inference methods for predictive machine learning with dependent data.
\newblock In \emph{Conference On Learning Theory}, pages 732--749. PMLR, 2018.

\bibitem[Cramer et~al.(2022)Cramer, Ray, Lopez, Bracher, Brennen, Castro~Rivadeneira, Gerding, Gneiting, House, Huang, et~al.]{cramer2022evaluation}
Estee~Y Cramer, Evan~L Ray, Velma~K Lopez, Johannes Bracher, Andrea Brennen, Alvaro~J Castro~Rivadeneira, Aaron Gerding, Tilmann Gneiting, Katie~H House, Yuxin Huang, et~al.
\newblock Evaluation of individual and ensemble probabilistic forecasts of covid-19 mortality in the united states.
\newblock \emph{Proceedings of the National Academy of Sciences}, 119\penalty0 (15):\penalty0 e2113561119, 2022.

\bibitem[Daniely et~al.(2015)Daniely, Gonen, and Shalev-Shwartz]{daniely2015strongly}
Amit Daniely, Alon Gonen, and Shai Shalev-Shwartz.
\newblock Strongly adaptive online learning.
\newblock In \emph{International Conference on Machine Learning}, pages 1405--1411. PMLR, 2015.

\bibitem[Dann et~al.(2023)Dann, Wei, and Zimmert]{dann2023best}
Christoph Dann, Chen-Yu Wei, and Julian Zimmert.
\newblock Best of both worlds policy optimization.
\newblock In \emph{International Conference on Machine Learning}, pages 6968--7008. PMLR, 2023.

\bibitem[Deng et~al.(2009)Deng, Dong, Socher, Li, Li, and Fei-Fei]{deng2009imagenet}
Jia Deng, Wei Dong, Richard Socher, Li-Jia Li, Kai Li, and Li~Fei-Fei.
\newblock Imagenet: A large-scale hierarchical image database.
\newblock In \emph{2009 IEEE conference on computer vision and pattern recognition}, pages 248--255, 2009.

\bibitem[Feldman et~al.(2021)Feldman, Bates, and Romano]{feldman2021improving}
Shai Feldman, Stephen Bates, and Yaniv Romano.
\newblock Improving conditional coverage via orthogonal quantile regression.
\newblock In \emph{Advances in Neural Information Processing Systems}, 2021.

\bibitem[Feldman et~al.(2023)Feldman, Ringel, Bates, and Romano]{feldman2023achieving}
Shai Feldman, Liran Ringel, Stephen Bates, and Yaniv Romano.
\newblock Achieving risk control in online learning settings.
\newblock \emph{Transactions on Machine Learning Research}, 2023.

\bibitem[Foreman-Mackey et~al.(2017)Foreman-Mackey, Agol, Ambikasaran, and Angus]{foreman2017fast}
Daniel Foreman-Mackey, Eric Agol, Sivaram Ambikasaran, and Ruth Angus.
\newblock Fast and scalable {G}aussian process modeling with applications to astronomical time series.
\newblock \emph{The Astronomical Journal}, 154\penalty0 (6):\penalty0 220, 2017.

\bibitem[Foster and Vohra(1998)]{foster1998asymptotic}
Dean~P Foster and Rakesh~V Vohra.
\newblock Asymptotic calibration.
\newblock \emph{Biometrika}, 85\penalty0 (2):\penalty0 379--390, 1998.

\bibitem[Gibbs and Candes(2021)]{gibbs2021adaptive}
Isaac Gibbs and Emmanuel Candes.
\newblock Adaptive conformal inference under distribution shift.
\newblock In M.~Ranzato, A.~Beygelzimer, Y.~Dauphin, P.S. Liang, and J.~Wortman Vaughan, editors, \emph{Advances in Neural Information Processing Systems}, volume~34, pages 1660--1672. Curran Associates, Inc., 2021.

\bibitem[Gibbs and Cand{\`e}s(2022)]{gibbs2022conformal}
Isaac Gibbs and Emmanuel Cand{\`e}s.
\newblock Conformal inference for online prediction with arbitrary distribution shifts.
\newblock \emph{arXiv preprint arXiv:2208.08401}, 2022.

\bibitem[Gradu et~al.(2023)Gradu, Hazan, and Minasyan]{gradu2023adaptive}
Paula Gradu, Elad Hazan, and Edgar Minasyan.
\newblock Adaptive regret for control of time-varying dynamics.
\newblock In \emph{Learning for Dynamics and Control Conference}, pages 560--572. PMLR, 2023.

\bibitem[Harries et~al.(1999)Harries, Wales, et~al.]{harries1999splice}
Michael Harries, New~South Wales, et~al.
\newblock Splice-2 comparative evaluation: Electricity pricing.
\newblock 1999.

\bibitem[Jin et~al.(2021)Jin, Huang, and Luo]{jin2021best}
Tiancheng Jin, Longbo Huang, and Haipeng Luo.
\newblock The best of both worlds: stochastic and adversarial episodic {MDP}s with unknown transition.
\newblock \emph{Advances in Neural Information Processing Systems}, 34:\penalty0 20491--20502, 2021.

\bibitem[Koenker and Bassett~Jr(1978)]{koenker1978regression}
Roger Koenker and Gilbert Bassett~Jr.
\newblock Regression quantiles.
\newblock \emph{Econometrica: Journal of the {E}conometric {S}ociety}, 46\penalty0 (1):\penalty0 33--50, 1978.

\bibitem[Koolen et~al.(2016)Koolen, Gr{\"u}nwald, and Van~Erven]{koolen2016combining}
Wouter~M Koolen, Peter Gr{\"u}nwald, and Tim Van~Erven.
\newblock Combining adversarial guarantees and stochastic fast rates in online learning.
\newblock \emph{Advances in Neural Information Processing Systems}, 29, 2016.

\bibitem[Lin et~al.(2022)Lin, Trivedi, and Sun]{lin2022conformal}
Zhen Lin, Shubhendu Trivedi, and Jimeng Sun.
\newblock Conformal prediction with temporal quantile adjustments.
\newblock \emph{Advances in Neural Information Processing Systems}, 35:\penalty0 31017--31030, 2022.

\bibitem[Lindemann et~al.(2023)Lindemann, Cleaveland, Shim, and Pappas]{lindemann2023safe}
Lars Lindemann, Matthew Cleaveland, Gihyun Shim, and George~J Pappas.
\newblock Safe planning in dynamic environments using conformal prediction.
\newblock \emph{IEEE Robotics and Automation Letters}, 2023.

\bibitem[Mykland(2003)]{mykland2003financial}
Per~Aslak Mykland.
\newblock Financial options and statistical prediction intervals.
\newblock \emph{The Annals of Statistics}, 31\penalty0 (5):\penalty0 1413--1438, 2003.

\bibitem[Noarov et~al.(2023)Noarov, Ramalingam, Roth, and Xie]{noarov2023high}
Georgy Noarov, Ramya Ramalingam, Aaron Roth, and Stephan Xie.
\newblock High-dimensional unbiased prediction for sequential decision making.
\newblock In \emph{OPT 2023: Optimization for Machine Learning}, 2023.

\bibitem[Robinson(1978)]{robinson1978sequential}
JA~Robinson.
\newblock Sequential choice of an optimal dose: A prediction intervals approach.
\newblock \emph{Biometrika}, 65\penalty0 (1):\penalty0 75--78, 1978.

\bibitem[Tian et~al.(2022)Tian, Nordman, and Meeker]{tian2022methods}
Qinglong Tian, Daniel~J Nordman, and William~Q Meeker.
\newblock Methods to compute prediction intervals: A review and new results.
\newblock \emph{Statistical Science}, 37\penalty0 (4):\penalty0 580--597, 2022.

\bibitem[Vovk(2002)]{vovk2002line}
Vladimir Vovk.
\newblock On-line confidence machines are well-calibrated.
\newblock In \emph{The 43rd Annual IEEE Symposium on Foundations of Computer Science}, pages 187--196. IEEE, 2002.

\bibitem[Vovk et~al.(2005)Vovk, Gammerman, and Shafer]{vovk2005algorithmic}
Vladimir Vovk, Alex Gammerman, and Glenn Shafer.
\newblock \emph{{Algorithmic Learning in a Random World}}.
\newblock Springer, 2005.
\newblock \doi{10.1007/b106715.}

\bibitem[Xu and Xie(2021)]{xu2021conformal}
Chen Xu and Yao Xie.
\newblock Conformal prediction interval for dynamic time-series.
\newblock In \emph{International Conference on Machine Learning}, pages 11559--11569. PMLR, 2021.

\bibitem[Xu and Xie(2023)]{xu2023sequential}
Chen Xu and Yao Xie.
\newblock Sequential predictive conformal inference for time series.
\newblock In \emph{International Conference on Machine Learning}, pages 38707--38727. PMLR, 2023.

\bibitem[Zaffran et~al.(2022)Zaffran, F{\'e}ron, Goude, Josse, and Dieuleveut]{zaffran2022adaptive}
Margaux Zaffran, Olivier F{\'e}ron, Yannig Goude, Julie Josse, and Aymeric Dieuleveut.
\newblock Adaptive conformal predictions for time series.
\newblock In \emph{International Conference on Machine Learning}, pages 25834--25866. PMLR, 2022.

\bibitem[Zimmert and Seldin(2021)]{zimmert2021tsallis}
Julian Zimmert and Yevgeny Seldin.
\newblock Tsallis-inf: An optimal algorithm for stochastic and adversarial bandits.
\newblock \emph{Journal of Machine Learning Research}, 22\penalty0 (28):\penalty0 1--49, 2021.

\bibitem[Zinkevich(2003)]{zinkevich2003online}
Martin Zinkevich.
\newblock Online convex programming and generalized infinitesimal gradient ascent.
\newblock In \emph{Proceedings of the 20th international conference on machine learning (icml-03)}, pages 928--936, 2003.

\end{thebibliography}
\appendix
\section{Additional proofs}

\subsection{Proof of Lemma~\ref{lem:qt_bound}}
The proof of this result is similar to the proof of Lemma 4.1 of \citet{gibbs2021adaptive}. We prove this by induction. First, $q_1\in[0,B]$ by assumption, so~\eqref{eqn:qt_bounded_range} is satisfied at time $t=1$. Next fix any $t\geq 1$ and assume $q_t$ lies in the range specified in~\eqref{eqn:qt_bounded_range}, and consider $q_{t+1}$. We now split into cases:
\begin{itemize}
    \item If $q_t\in [0,B]$, then we have
    \[q_{t+1} = q_t + \eta_t \left(\one{Y_t\not\in\cC_t(X_t)}-\alpha\right) \\\in [q_t - \eta_t \alpha, q_t + \eta_t (1-\alpha)]\\  \subseteq [-\alpha M_t, B + (1-\alpha)M_t].\]
    \item  If $q_t\in (B, B+(1-\alpha)M_{t-1}]$, then we must have $\cC_t(X_t) = \cY$. Then $\one{Y_t\not\in\cC_t(X_t)}=0$, and so
    \[q_{t+1}  = q_t - \eta_t \alpha \in  [B-\eta_t\alpha , B + (1-\alpha)M_{t-1}]\subseteq  [-\alpha M_t, B + (1-\alpha)M_t].\]
    \item If $q_t\in[-\alpha M_{t-1},0)$, then we must have $\cC_t(X_t) = \emptyset$. Then $\one{Y_t\not\in\cC_t(X_t)}=1$, and so
    \[q_{t+1} =  q_t +\eta_t (1-\alpha) \in [-\alpha M_{t-1},\eta_t(1-\alpha)]\subseteq [-\alpha M_t,B+(1-\alpha) M_t].\]
\end{itemize}
In all cases, then,~\eqref{eqn:qt_bounded_range} holds for $t+1$ in place of $t$, which completes the proof.

\subsection{Proof of~\ref{eqn:martingale}}
We need to prove that $Z_t$ converges almost surely (note that the limit of $Z_t$ may be a random variable).
For each $t\geq 1$, we have
\[\PPst{Y_t\in\cC_t(X_t)}{\{(X_r,Y_r)\}_{r<t}}\\
= \PPst{s_t(X_t,Y_t)\leq q_t}{\{(X_r,Y_r)\}_{r<t}} = \mathsf{Coverage}_t(q_t),\]
since $q_t$ and $s_t$ are functions of $\{(X_r,Y_r)\}_{r<t}$ and are therefore independent of $(X_t,Y_t)\sim P$. This proves that $Z_t$ is a martingale with respect to the filtration generated by the sequence of data points.
We also have $\sup_{t\geq 1}\textnormal{Var}(Z_t)< \infty$, since we have assumed $\sum_{t=1}^\infty\eta_t^2<\infty$.
This means that $Z_t$ is a uniformly integrable martingale, and therefore, $Z_t$ converges almost surely (to some random variable), by Doob's second martingale convergence theorem.

\subsection{Proofs of Corollaries~\ref{cor:cor1} and~\ref{cor:cor2}}
As for the theorems, it suffices to prove Corollary~\ref{cor:cor2}, since Corollary~\ref{cor:cor1} is simply a special case.

Using the notation defined in the proof of Theorem~\ref{thm:q_converge2}, suppose that events $\mathcal{E}_Z$ and $\mathcal{E}_s$ both hold. Now we need to show that $\mathsf{Coverage}_t(q_t)\rightarrow 1-\alpha$ holds as well. Fix any $\eps>0$. Since $s(X,Y)$ has a continuous distribution, the map $q\mapsto \mathsf{Coverage}(q)$ is continuous, and so we can find some $\delta>0$ such that
\[\left|\mathsf{Coverage}(q) - \mathsf{Coverage}(q^*)\right| \leq \eps/2\textnormal{ for all }q\in q^*\pm \delta.\] 
Moreover, $\mathsf{Coverage}(q^*)=1-\alpha$, since the distribution of $s(X,Y)$ is continuous and $q^*$ is its $(1-\alpha)$-quantile, so we have
\[\left|\mathsf{Coverage}(q) - (1-\alpha)\right| \leq \eps/2\textnormal{ for all }q\in q^*\pm \delta.\] 
Next, by Theorem~\ref{thm:q_converge2}, for all sufficiently large $t$, we have
\[|q_t - q^*| \leq \delta.\]
By definition of the event $\mathcal{E}_s$, for all sufficiently large $t$ we have
\[\left|\mathsf{Coverage}_t(q^*-\delta) - \mathsf{Coverage}(q^*-\delta)\right| \leq \eps/2\]
and
\[\left|\mathsf{Coverage}_t(q^*+\delta) - \mathsf{Coverage}(q^*+\delta)\right| \leq \eps/2.\]
Then, combining all of these calculations, for all sufficiently large $t$ we have
\[
    \mathsf{Coverage}_t(q_t)
    \geq \mathsf{Coverage}_t(q^* - \delta)
    \geq \mathsf{Coverage}(q^*-\delta) - \eps/2
    \geq (1-\alpha -\eps/2) - \eps/2
    =1-\alpha-\eps,
\]
where the first step holds since $q_t\geq q^*-\delta$, and $q\mapsto \textnormal{Coverage}_t(q)$ is nondecreasing. Similarly, for all sufficiently large $t$ it holds that
\[\mathsf{Coverage}_t(q_t) \leq 1-\alpha+\eps.\]
Since $\eps>0$ is arbitrary, this completes the proof.

\subsection{Proof of Proposition~\ref{prop:converge_rate}}
By Lemma~\ref{lem:qt_bound}, $q_t \in[-\alpha c, B + (1-\alpha)c]$ for all $t$ (since $\eta_t\leq c$ for all $t$). Since $q^*\in[0,B]$, we therefore have $|q_t - q^*|\leq B+c$ almost surely for all $t$. We also have $2\gamma\delta\leq 1$ and $\delta \leq B \leq B+c$, since the density of $s(X,Y)$ is supported on $[0,B]$ and must integrate to $1$.

Next, by the assumptions of the proposition, for any $q_t\geq q^*$, if $q_t\leq q^*+\delta$ then
\[\mathsf{Coverage}(q_t) \geq 1-\alpha + \gamma(q_t - q^*)\]
while if $q_t>q^*+\delta$ then
\[\mathsf{Coverage}(q_t) \geq \mathsf{Coverage}(q^*+\delta)\geq 1-\alpha + \gamma\delta.\]
Either way, then, if $q_t\geq q^*$ then
\[\mathsf{Coverage}(q_t) \geq (1-\alpha) + (q_t-q^*) \cdot \frac{\gamma\delta}{B+c}.\]
A similar calculations shows that if $q_t\leq q^*$, then
\[\mathsf{Coverage}(q_t) \leq (1-\alpha) - (q^*-q_t) \cdot \frac{\gamma\delta}{B+c}.\]
Defining $a = \frac{\gamma\delta}{B+c}>0$,
we therefore have
\begin{equation}\label{eqn:coverage_ratio}\frac{\mathsf{Coverage}(q_t) - (1-\alpha)}{q_t - q^*} \geq a\end{equation}
whenever $q_t \neq q^*$. Note that we must have $a \leq 1/c$, by construction.

Next, from the update step~\eqref{eq:q_update}, we have 
\[q_{t+1} - q^* = (q_t - q^*) + \eta_t \big((1-\alpha) - \one{Y_t\in\cC_t(X_t)}\big).\]
Since $\PPst{Y_t\in\cC_t(X_t)}{\{(X_r,Y_r)\}_{r<t}} = \mathsf{Coverage}(q_t)$, we then calculate
\[\EEst{q_{t+1} - q^*}{\{(X_r,Y_r)\}_{r<t}} = (q_t - q^*) + \eta_t \big((1-\alpha) -\mathsf{Coverage}(q_t)\big),\]
and
\[\textnormal{Var}\big(q_{t+1}-q^*\mid\{(X_r,Y_r)\}_{r<t}\big) = \eta_t^2 \cdot  \mathsf{Coverage}(q_t) \cdot (1-\mathsf{Coverage}(q_t)) \leq \eta_t^2/4.\]
Therefore,
\begin{multline*}\EEst{(q_{t+1}-q^*)^2}{\{(X_r,Y_r)\}_{r<t}} \leq 
\left((q_t - q^*) + \eta_t \big((1-\alpha) -\mathsf{Coverage}(q_t)\big)\right)^2 + \eta_t^2/4\\
\leq (q_t-q^*)^2 \cdot (1-a\eta_t)^2 + \eta_t^2/4,\end{multline*}
where the last step holds by~\eqref{eqn:coverage_ratio} above. After marginalizing, then,
\[\EE{(q_{t+1}-q^*)^2} \leq \EE{(q_t-q^*)^2} \cdot (1-a\eta_t)^2 + \eta_t^2/4.\]
Next recall $\eta_t = ct^{-1/2-\eps}$ for each $t$. Fix some $T\geq 1$ that satisfies $T^{1/2-\eps} \geq \frac{1/2 + \eps}{ac}$.  First, since $|q_t-q^*| \leq B+c$ for all $t$ as above, by choosing $b\geq (B+c)^2T^{1/2+\eps}$  we must have $(q_t-q^*)^2\leq bt^{-1/2-\eps}$ for all $t\leq T$, almost surely. Next, for each $t\geq T$, we proceed by induction. Assume $\EE{(q_t - q^*)^2}\leq bt^{-1/2-\eps}$. Then
\begin{align*}
    \EE{(q_{t+1}-q^*)^2} 
    &\leq \EE{(q_t-q^*)^2} \cdot (1-a\eta_t)^2 + \eta_t^2/4\\
    &=\EE{(q_t-q^*)^2}(1-2a\eta_t) + \left(\EE{(q_t-q^*)^2}\cdot a^2+ 1/4\right)\eta_t^2 \\
    &\leq bt^{-1/2-\eps}\left( 1-2act^{-1/2-\eps}\right) + ((B+c)^2a^2+1/4)c^2 t^{-1-2\eps}\\
    &=bt^{-1/2-\eps} -2abc t^{-1-2\eps}+ ((B+c)^2a^2+1/4)c^2 t^{-1-2\eps}\\
    &\leq b\left(t^{-1/2-\eps} -ac t^{-1-2\eps}\right),
\end{align*}
where the last step holds as long as we choose $b\geq \frac{((B+c)^2a^2+1/4) c}{a}$. And, since $t\geq T$, we have 
\[act^{-1-2\eps} = act^{1/2-\eps} \cdot t^{-3/2-\eps} \geq  acT^{1/2-\eps} \cdot t^{-3/2-\eps} \geq (1/2+\eps)t^{-3/2-\eps} \geq t^{-1/2-\eps} - (t+1)^{-1/2-\eps},\]
where the last step holds since $t\mapsto t^{-1/2-\eps}$ is convex, with derivative $-(1/2+\eps)t^{-3/2-\eps}$. Therefore, we have verified that $\EE{(q_{t+1}-q^*)^2} \leq b(t+1)^{-1/2-\eps}$, as desired.

\section{Additional experiments}
\begin{figure}[t]
    \centering
    \includegraphics[width=\textwidth]{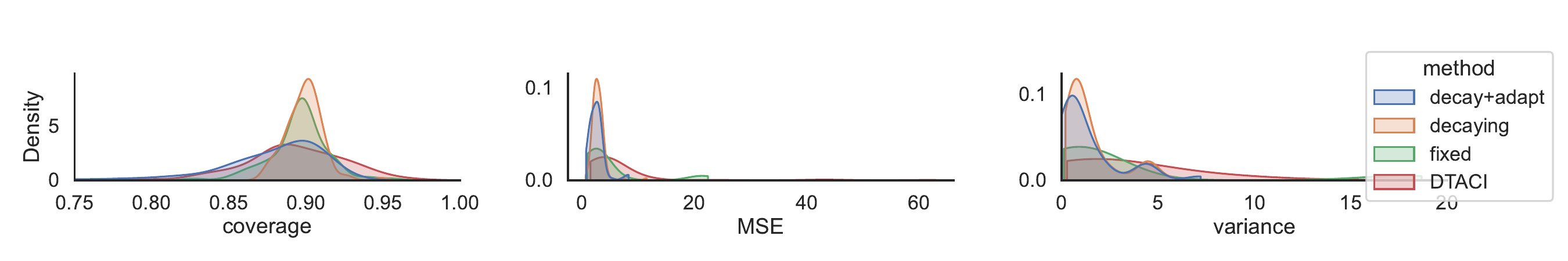}
    \caption{\textbf{Density plots of results on M4 datasets.} These plots show the same quantities as in Table~\ref{tab:M4}, but now as histograms over the time-series in M4.}
    \label{fig:M4}
\end{figure}
We compare against two additional methods: first, ``decay+adapt'', a variant of our procedure that decays until it detects a change point, then resets the learning rate. Change points are identified when at least $N_{\rm miscoverage}$ consecutive miscoverage events or $N_{\rm coverage}$ events are observed in a row (we set these constants to 10 and 30 by default, respectively).
When a change point is identified, the learning rate is reset to $\frac{\hat{B}}{(t-T_{\rm changepoint})^{1/2 + \epsilon}}$, where $T_{\rm changepoint}$ is the time at which the changepoint is detected and $\epsilon \in (0,1/2)$.
In these experiments, like in the main text, we set $\epsilon = 0.1$.

We additionally compare against DTACI~\cite{gibbs2022conformal}, an adaptive-learning-rate variant of ACI that uses multiplicative weights to perform the updates (see~\cite{gibbs2022conformal} for further details.)

We compare these methods on a dataset of over 3000 time series subsampled from the M4 time series dataset.
This dataset is a diverse array of time series with varying numbers of samples and distribution shifts.
Code is available in our GitHub repository to run on all 100,000 time series in M4; here, we show results on the first 3000.

\begin{figure}
    \centering
    \includegraphics[width=0.4\textwidth]{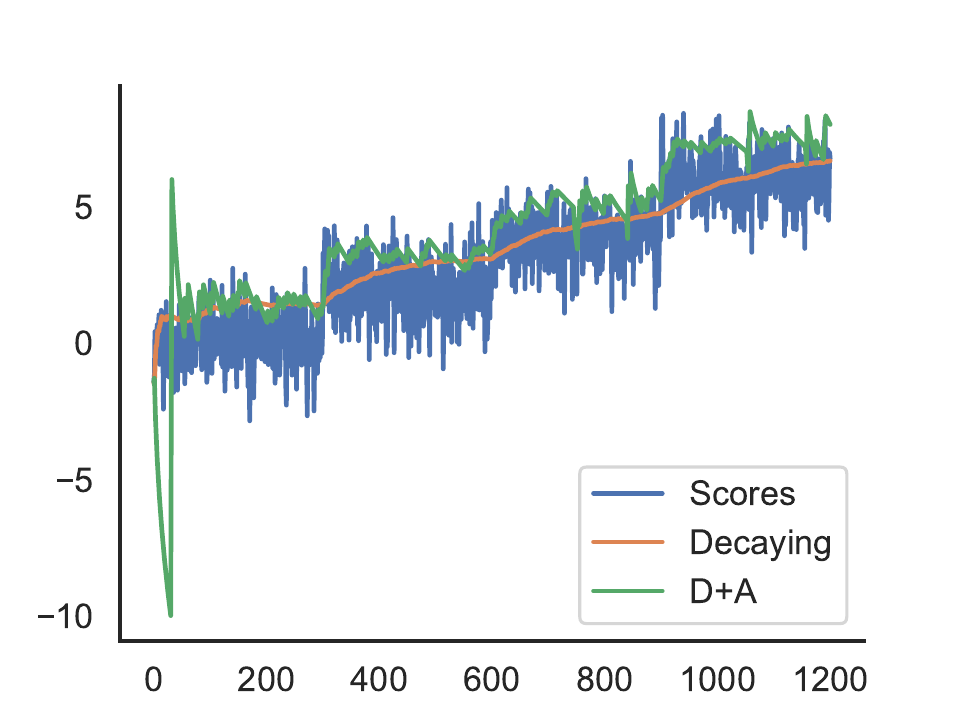}
    \caption{\textbf{Simulation comparison of decaying step size and `decay+adapt'.} The raw score sequence is shown in blue, the decaying step size sequence is in orange, and `decay+adapt' is in green.}
    \label{fig:simulation}
\end{figure}

Finally, to showcase the conceptual differences between the standard decaying learning rate sequence and the `decay+adapt' method, we display a simulated score sequence in Figure~\ref{fig:simulation}.
Here, the scores are simulated from $\mathcal{N}(\mu_t, 1)$, where $\mu_t = 0$ for the first thousand time steps, $\mu_t = 2$ for the second thousand, $\mu_t = 4$ for the third thousand, and $\mu_t = 6$ for the final thousand.
Especially towards the end of the time series, `decay+adapt' can more quickly adjust to the change points.
\begin{table}[t]
    \centering
    \begin{tabular}{lrrrr}
    \toprule
    method & coverage & variance & MSE & infinite sets \\
    \midrule
    DTACI & 0.895491 & 4.107740 & 5.439935 & 0.036584 \\
    decay+adapt & 0.885174 & 1.144337 & 1.967552 & 0.005011 \\
    decaying & 0.900495 & 1.320579 & 2.366297 & 0.006883 \\
    fixed & 0.901636 & 1.580243 & 2.922989 & 0.011179 \\
    \bottomrule
    \end{tabular}
    \caption{\textbf{Table of results on M4 datasets.} The table shows average results over all time series in the dataset---thus, all columns should be interpreted on average \emph{over time-series in M4}. The coverage column displays the long-run coverage. The variance column shows the variance of the quantile normalized by the variance of the score sequence. The MSE column shows the squared error of the quantile normalized by the variance of the score sequence. Finally, the infinite sets column shows the fraction of time steps in the sequence for which the output is an infinite-width prediction set.}
    \label{tab:M4}
\end{table}

\end{document}